# SPICE-HL³: Single-Photon, Inertial, and Stereo Camera dataset for Exploration of High-Latitude Lunar Landscapes

David Rodríguez-Martínez[1,*], Dave van der Meer[2], Junlin Song[2], Abhishek Bera[2], C.J. Pérez-del-Pulgar[1], and Miguel Angel Olivares-Mendez[2]
*corresponding author: david.rm@uma.es

[1]Space Robotics Lab, Department of Systems Engineering and Automation, University of Malaga, Spain
[2]Space Robotics Research Group, Interdisciplinary Research Center for Security, Reliability, and Trust (SnT), University of Luxembourg, Luxembourg

**Abstract**

Exploring high-latitude lunar regions presents a challenging visual environment for robots. The low sunlight elevation angle and minimal light scattering result in a visual field dominated by a strong contrast featuring long, dynamic shadows. Reproducing these conditions on Earth requires sophisticated simulators and specialized facilities. We introduce a unique dataset recorded at the LunaLab from the SnT - University of Luxembourg, an indoor test facility designed to replicate the optical characteristics of multiple lunar latitudes. Our dataset includes images, inertial measurements, and wheel odometry data from robots navigating different trajectories under multiple illumination scenarios, simulating high-latitude lunar conditions from dawn to nighttime with and without the aid of headlights, resulting in 88 distinct sequences containing a total of 1.3M images. Data was captured using a stereo RGB-inertial sensor, a monocular monochrome camera, and, for the first time, a novel single-photon avalanche diode (SPAD) camera. We recorded both static and dynamic image sequences, with robots navigating at slow (5 cm/s) and fast (50 cm/s) speeds. All data is calibrated, synchronized, and timestamped, providing a valuable resource for validating perception tasks from vision-based autonomous navigation to scientific imaging for future lunar missions targeting high-latitude regions or those intended for robots operating across perceptually degraded environments.

## Background & Summary

A great number of planetary exploration and prospecting missions are currently aiming at the lunar poles due to their expected abundance of resources and their potential to address key questions about the Moon's origin and evolution[1,2,3]. However, high-latitude lunar regions pose significant challenges for robotic and perception systems. The lunar South Pole, in particular, presents one of the most perceptually challenging environments any planetary rover has ever faced while featuring some of the Moon's most extreme terrain elevation changes. The small lunar obliquity to the ecliptic (of just 1.54◦) produces very low solar elevation angles at the poles (often <10º [4]), which, combined with the rugged terrain, results in a visual field dominated by dim lights and long, dynamic shadows. These shadows sweep across a monochromatic and, at times, featureless environment where negligible light scattering leads to a stark contrast between the bright highland regolith and the darkness of crater floors and permanently shadowed areas.

Robots must, therefore, move fast in this environment—faster than ever before[5]—to avoid getting stranded in the moving shadows or succumbing to the freezing temperatures in permanent cold traps. Regardless of their level of automation, onboard perception systems must be capable of effectively resolving landmarks and other surrounding features with a high enough signal-to-noise to inform subsequent actions, even in darkness. As a result, successful navigation and scientific characterization in high-latitude lunar regions requires high frame rates, rapid exposure adaptations, and high dynamic range (HDR) imaging capabilities, which may demand the use of innovative sensing technologies and data processing algorithms for localization, mapping, and science-driven tasks.

Opportunities to acquire in-situ data to inform planetary exploration missions are exceptionally scarce. On Earth, replicating the visual conditions of places like the lunar poles to a high degree of fidelity becomes exceedingly difficult. With this dataset, we aim to bridge this gap. SPICE-HL³ ([6]) presents proprio- and exteroceptive data recorded with multiple fictionlab's Leo rovers at the LunaLab of the Space Robotics Group at the SnT - University of Luxembourg[7]. Our contribution is unique in two ways. First, we make public the most exhaustive real lunar-like robotic onboard data to date, collected through a series of trajectories designed to mimic the negotiation of obstacles and navigation patterns of a rover traversing across high-latitude lunar regions. Visual data contained most perceptual artifacts expected from these missions, namely harsh and dynamic shadows, lens flares, motion blur, accumulated dust, Poissonian noise, and HDR, resulting in over- and undersaturated pixel values. Second, and for the first time, it includes data recorded using a novel Single-Photon Avalanche Diode (SPAD) camera. As we will discuss in the Method section, SPAD cameras represent a new sensing paradigm for vision-based robotic systems, making SPADs promising candidates as primary sensing instruments for exploring perceptually degraded environments like the lunar poles. The acquired data can be used to test and validate vision-driven tasks, from navigation approaches to sensor processing algorithms. For more information about how the dataset can be use, in a separate publication[8], we conducted a preliminary evaluation of the imaging and task-driven performance (scene segmentation, lander identification, and rover localization) of the SPAD with respect to that of a conventional monochromatic camera using data extracted from this dataset.

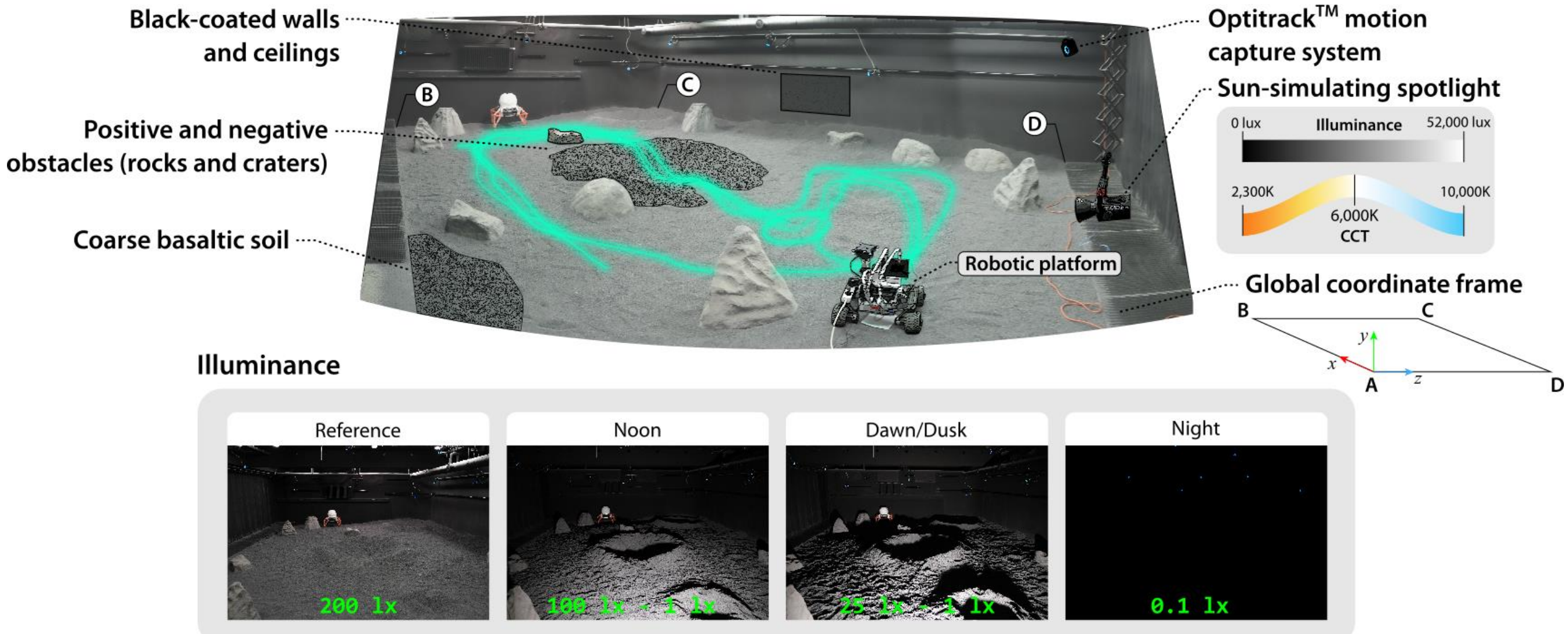

Fig. 1: The LunaLab facility at the University of Luxembourg is designed to simulate lunar photometric conditions across multiple latitudes. This dataset contains images captured under four different illumination conditions, namely: reference, noon, dawn/dusk, and nighttime. Illuminance values were measured with a luxemeter located in bright and shaded areas across the center of the test field. The origin of the global coordinate frame is located at the camera position from where the top image was taken ('A'); the other 3 vertices of the global plane are also denoted in the image ('B', 'C', and 'D'). A visual reference of all the trajectories followed is overlaid in the image.

*Related datasets*

Devising and testing new algorithms for vision-driven tasks require empirical data from higher lunar latitudes or representative datasets, which are currently lacking. Synthetic simulators offer cost-efficient alternatives[9, 10, 11], but they struggle to capture the complex interplay of photometric, material, and sensor characteristics that occur in real or analog environments; e.g., the effect of the albedo and the scattering of light through the lunar regolith are challenging to accurately and efficiently model in synthetic environments, particularly when intended for large-scale robotic simulations.

In-situ data such as the RCLLD dataset[12], acquired by the Chang'e-3, Chang'e-4, and Chang'e-5 lunar landers, as well as panoramic images captured by the Yutu-1 and Yutu-2 lunar rovers, provide valuable labeled imagery across diverse lighting conditions but lack the temporal continuity and ground-truth data required when developing and validating new algorithms.

In contrast, multiple published datasets attempt to recreate the unstructured and extreme nature of planetary environments through data recorded in analog Mars- and lunar-like regions on Earth. These include the Devon Island dataset[13] recorded in the Canadian High Arctic, the LRNT[14] and S3LI[15] datasets captured at Mount Etna by the German Aerospace Center (DLR), or the BASEPROD dataset[16] recorded in the natural badlands of Bardenas Reales, in northern Spain. While these provide ground-truth measurements and annotated data, they fail to capture the extreme and unique illumination conditions of high-latitude lunar regions (i.e., polar Sun elevation angles, milli-lux illumination, negligible light scattering): a gap we strove to bridge with SPICE-HL$^3$ ([6]).

Only a handful of facilities can replicate the complex lunar polar photometric conditions. To our knowledge, just two publicly available datasets attempt to reproduce such conditions: NASA Ames' POLAR Stereo[17] and POLAR Traverse[18] datasets. Both feature stereo imagery of regolith-filled test beds (JSC-1A and modified LHS-1, respectively) scattered with rocks and craters under 1kW low-angle tungsten-halogen lighting, alongside LiDAR scans and radiometric calibration. The POLAR Stereo dataset provides static viewpoints of varied landscape configurations, while the POLAR Traverse dataset includes simulated motion sequences with images taken at 1 m intervals. Despite the rigorousness in mimicking the photometric conditions expected in high-latitude lunar regions, the static nature of the data, sparse stereo intervals, limited linear trajectories, and the fact that background elements (walls and equipment) are visible in some of the images, reduce their fidelity and hinder their use for navigation research. SPICE-HL$^3$ ([6]) overcomes these limitations by providing both static and dynamic sequences at varied speeds (5 and 50 cm/s), compound trajectories with higher frame rates, controlled lighting variations, and minimized background artifacts. Table I outlines the key characteristics of these various datasets in comparison with ours.

TABLE I: Comparison of lunar datasets with SPICE-HL$^3$ ([6]).

| Dataset | Environment | Temporal Continuity | Dynamic Sequences | GT | Polar Lighting | Modalities |
|---|---|---|---|---|---|---|
| ALLD[9] | Synthetic | ✗ | ✗ | ✗ | ✓ | Mono, Labeled |
| RCLLD[12] | Real | ✗ | ✗ | ✗ | ✓ | Mono, Labeled |
| Devon Island[13] | Analog | ✓ | ✓ | ✓ | ✗ | Stereo, Inclinometer, Sun tracker |
| LRNT[14] | Analog | ✓ | ✓ | ✓ | ✗ | Stereo, IMU, Wheel Odometry |
| S3LI[15] | Analog | ✓ | ✓ | ✓ | ✗ | Stereo, LiDAR, IMU |
| BASEPROD[16] | Analog | ✓ | ✓ | ✓ | ✗ | Stereo, Thermal, IMU, Force/Torque |
| POLAR Stereo[17] | Analog | ✗ | ✗ | ✓ | ✓ | Stereo, LiDAR |
| POLAR Traverse[18] | Analog | ✓ | ✗ | ✓ | ✓ | Stereo, LiDAR |
| **SPICE-HL$^3$ ([6]) (ours)** | Analog | ✓ | ✓ | ✓ | ✓ | Mono, Stereo, SPAD, IMU, Wheel Odometry |

GT:Ground Truth

## METHODS

### *Testing facility*

The LunaLab consists of an 11 × 8 m$^2$ lunar analog terrain designed to mimic the optical conditions found across different lunar latitudes (see Fig. 1). The facility is equipped with an Aputure's 600W LightStorm 600c Pro spotlight, whose vertical and horizontal position along the shorter side of the test bed, its color temperature (2,300 K–10,000 K), and its luminance (69 lux–51,100 lux) can be fully adjusted to simulated different sunlight elevation angles and daytime conditions. We recorded data under four different illumination conditions (Reference, Noon, Dawn/Dusk, and Night), as illustrated in Fig. 1. Noon and Dawn/Dusk conditions were obtained by positioning the spotlight at a height of 80 cm and 40 cm, respectively; simulating the low range of Sun elevation angles expected across a full lunar day at the poles (8º and 4º, respectively, with respect to the center of the test bed). For all the trajectories, we used a CCT of 6,000 K to simulate sunlight color temperature in space. The testbed is filled with 20 tons of coarse basaltic gravel and features rocks and craters of varying sizes and shapes, which introduce additional shadows and occlusions. Black-coated walls and ceilings surround the test bed to minimize secondary light reflections, simulating the absence of a scattering atmosphere.

### *Sensor setup*

Our dataset was recorded using a variety of passive optics, including a monochromatic camera (Teledyne FLIR BlackflyS), a stereo-inertial sensor (StereoLabs ZED2), and a single-photon camera (PiImaging SPAD512)(refer to the Method section for additional details on single-photon imaging). The specifications of each of the cameras used are listed in Table II. Inertial measurements were recorded either through the built-in Inertial Measurement Unit (IMU) on the Leo rover or the one on the ZED2. Additionally, odometric data was measured through in-wheel 12-PPR incremental rotary encoders.

TABLE II: Camera sensors used and their specifications

| | BlackflyS | ZED2 | SPAD512 |
|---|---|---|---|
| Sensor size | 6.23×4.98 mm | 4.8×3.6 mm | 9.5×9.5 mm |
| Resolution | 720×540 | 640×360 | 512×512 |
| Pixel Pitch | 6.9 μm | 2 μm | 16.38 μm |
| Bit Depth | 8-bit | 8-bit | 1-bit |
| EFL | 14.24 mm f1.4 | 2.12 mm f2.0 | 14.88 mm f1.4 |
| FoV | 21.6° ×12.3° | 100° ×70° | 35.5° ×35.5° |
| Baseline | n/a | 120 mm | n/a |

EFL: Effective Focal Length; FoV: Field of View

The sensors used for data recording were mounted on two fictionlab's Leo rovers, each with a different sensor configuration. One rover incorporated the FLIR BlackflyS and the SPAD camera, while the ZED2 stereo camera rig was mounted on a second rover. Both rovers are powered by 4 Bühler DC motors controlled by a Raspberry Pi 4B. During data acquisition, the rovers were teleoperated via ROS2 from outside the lab over a 2.4 GHz Wifi network. Data for the FLIR and SPAD cameras were recorded on a separate Lenovo ThinkCentre M900 PC, while the stereo camera and IMU data were recorded directly on the rover's onboard computer. The FLIR and SPAD cameras were positioned approximately 470 mm from the ground, and their mounts were designed to maintain a constant 20° pitch, resulting in a look-ahead distance of 0.5–10 m, which allowed the cameras to resolve immediate obstacles as well as part of the horizon while avoiding most of the surrounding background elements. In the case of the ZED2 camera, the mount was positioned 550 mm from the ground and at a 25º pitch incline. Fig. 2 illustrates each rover configuration and sensor layout. The rovers also incorporated a headlight, which was used to record variations of illuminance during one of the trajectories (see the Data Record section for details on the different trajectories). Ground truth of the rovers' movement was acquired via 12 OptiTrack™ Prime$^x$ motion capture cameras with sub-millimeter accuracy.

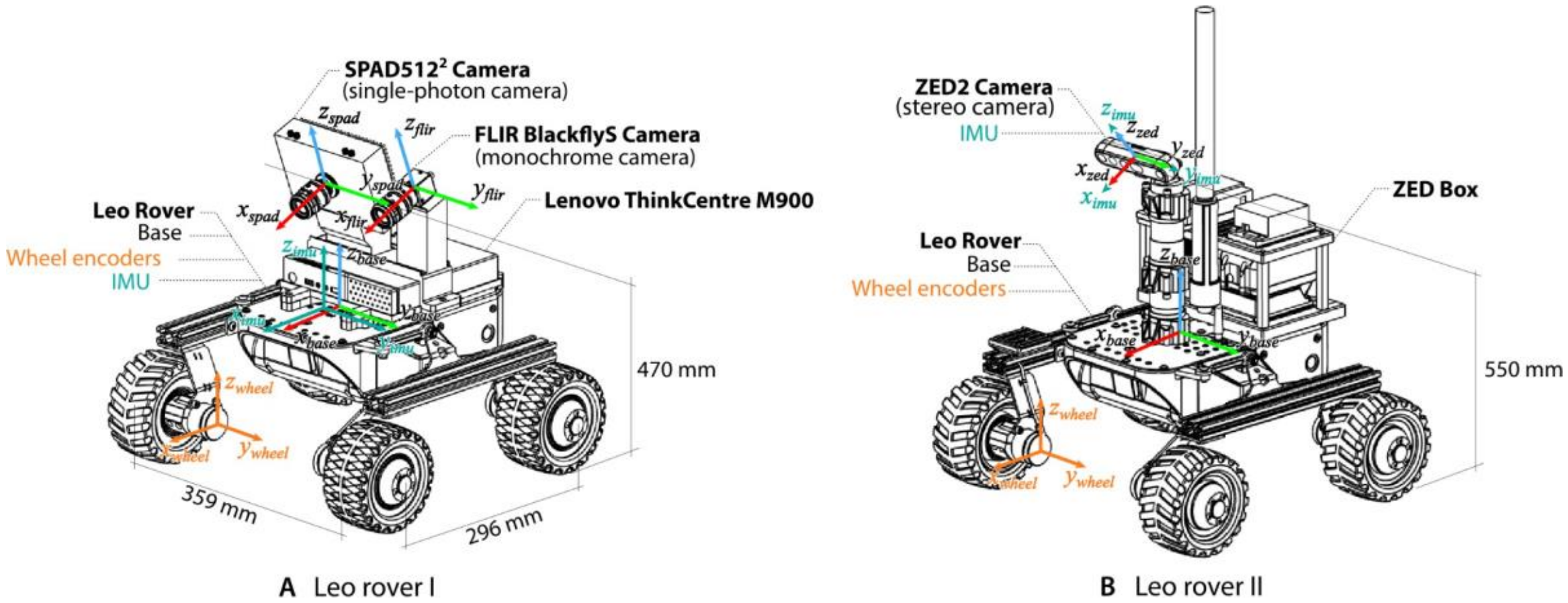

Fig. 2: Rovers configuration and sensor layout.

### *SPAD Camera*

In this dataset, and as far as we know, for the first time in a planetary exploration setting, we present data recorded with a Single-Photon Avalanche Diode (SPAD) camera. SPAD arrays are a novel set of imaging devices capable of detecting the arrival of individual photons between sub-10 picosecond time intervals, with wide dynamic ranges (>100 dB), and exceptionally low noise[19,20]. Single-photon resolution and minimal dead times are possible thanks to the use of Geiger-mode avalanche photodiodes and active quenching. Unlike conventional cameras, which provide a digitized signal equivalent to the amount of light intensity accumulated over time (0–255 per pixel, for 8-bit images), a SPAD pixel provides a direct binary output (0: no photon detected, 1: photon detected). Their extremely light-sensitive pixels, paired with fast data acquisition and their versatile data processing capabilities, make SPAD cameras a promising sensing technology for a wide range of applications, particularly those dealing with visually challenging scenarios where conventional cameras struggle to achieve enough signal-to-noise. SPADs have been primarily used in passive mode for biomedical imaging applications[21] or in active mode, when combined with a synchronized light source, for time-of-flight (ToF) and Light Detection and Ranging (LiDAR) applications[22].

In our case, we used the SPAD512 model from Pi Imaging Technologies as a passive camera to record monochromatic images meant to assist in visual-based scientific imaging and autonomous navigation pipelines. This camera has a resolution of 512×512 pixels, and it is capable of acquiring up to 100,000 binary frames per second (i.e., 1-bit images) with a minimum exposure time of 20 ns per frame. These binary frames could be later integrated to digitize images at higher bit depths. Fig. 3 illustrates examples of the raw binary frames output by the SPAD and their equivalent higher bit-depth integration. We have developed a series of scripts that facilitate the programmatic acquisition of binary frames from the SPAD512 for robotic applications (i.e, acquisition of a predefined batch of binary frames at a given batch rate, which is not provided natively by the camera firmware) and the subsequent frame integration (refer to the Code Availability section for details).

## DATA RECORD

The dataset is publicly available at Zenodo[6]. It includes measurements from 88 distinct sequences recorded along seven different trajectories (*A*–*G*). Each trajectory presents multiple time-synchronized monocular and stereo image sequences, inertial measurements, and wheel odometry data captured under different illumination conditions, with variations in the type of movements performed, the speed of the rover, and/or the type of cameras used. We intended to record as many variations in movement and illuminance as possible to replicate the set of expected operational conditions during actual exploration missions to the lunar poles, resulting in a total of 1,289,958 images. The characteristics of each of the trajectories are listed in Table III, and further details are provided in these section. The ground truth position of the rover with sub-millimeter accuracy is provided for every sequence. The layout of the test field and the rovers' configuration remain invariant throughout the data collection campaign. The ground was raked after each run to prevent footprints from creating unnatural features on the terrain.

### *Data format*

**Image files** from the SPAD512 and ZED2 cameras are provided in lossless PNG format, while images captured by the FLIR BlackflyS during *trajectory A* were recorded as JPG. The images captured by the FLIR and ZED2 cameras are stored in 8-bit format. While the FLIR frame rate varied with the preset exposure, the ZED2 stereo image sequences were acquired at 5 Hz for RGB and 10 Hz for monochromatic images. Disparity maps generated from the stereo images have been computed using Stereo Block Matching (SBM) and saved as 8-bit color-coded PNG files. The Python script used to generate disparity maps is provided in the associated GitHub repository (refer to the Code Availability section for details). In the case of the SPAD camera, we offer different formats based on the type of trajectory. For the static trajectory (*trajectory A*), we provide sequences of raw binary frames. In contrast, the dynamic trajectories include image sequences in a 4-bit format. This choice was due to memory constraints of the SPAD512, which limits continuous 1-bit-mode data capture to a maximum of 130,000 frames per acquisition. Given the exposure time used for dynamic sequences (typically between 0.5µs–25µs per binary frame), this constraint would have restricted the total recording time per trajectory to less than 3 s. Even though the exposure time could be manually adjusted, another limitation of the SPAD512 was its lack of built-in support for acquiring data at a given frame rate (i.e., with a set deadtime in between frames). To overcome these limitations, we wrote a series of custom scripts that allowed us to programmatically command the SPAD512 to capture an 8-bit-equivalent number of frames (255 in 1-bit mode, 15 in 4-bit mode) at a predefined frame rate (e.g., every 20 ms). Despite this improvement, we encountered a bottleneck during image acquisition in 1-bit mode that might be associated with network constraints or internal camera limitations. It took on average 100 ms for the SPAD camera to process each batch of 255 binary frames regardless of the exposure time, effectively limiting the frame rate in 1-bit mode to about 10 Hz. For this reason, we opted for continuous 4-bit mode acquisition in the case of dynamic motion sequences. All the associated scripts are included in the GitHub repository accompanying this dataset. We also provide each camera metadata—including *frame ID*, *timestamp*, *height*, *width*, and *file encoding*—organized into corresponding CSV text files. Example images from all the cameras and the different SPAD formats are illustrated in Fig. 4.

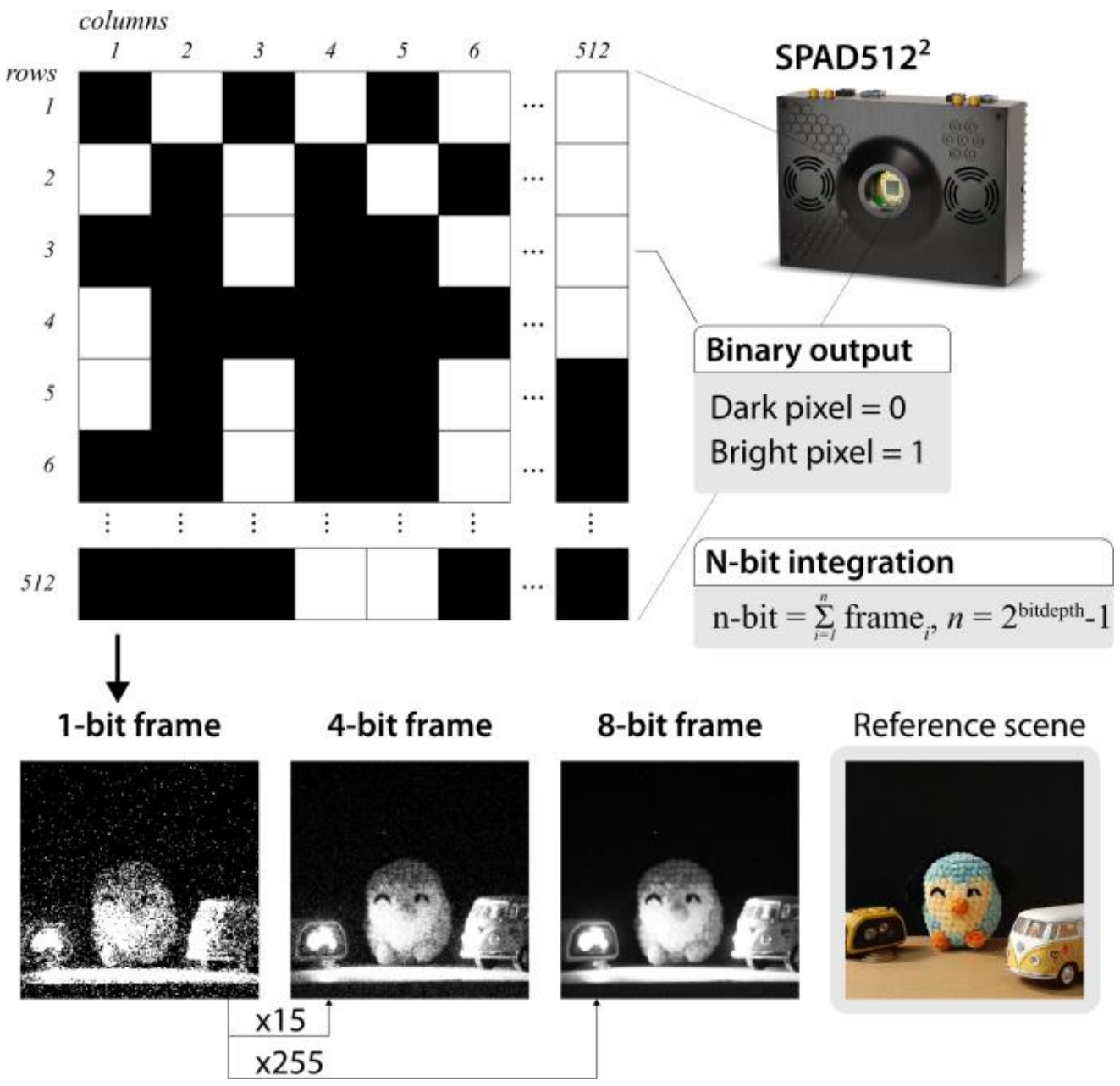

Fig. 3: The SPAD512 camera acquires raw binary frames, which are a set of 1-bit-per-pixel frames that could be later integrated to form tailored bit-depth images.

TABLE III: Description of every trajectory recorded as part of the SPICE-HL$^3$ dataset[6], their different visual conditions, and associated data. "Seq." indicates the number of distinct sequences per trajectory. Length for trajectories B–E is computed as the average among the different slow and fast runs. For these trajectories, *slow* / *fast* times are listed.

| Trajectories | Seq. | Description | Imaging Mode | Illumination Mode | Rover Speed | Cameras | Headlights | Length | Duration |
|---|---|---|---|---|---|---|---|---|---|
| Trajectory A | 70 | Compound static | Static | All | n/a | SPAD & FLIR | On/Off | 4.37m | 82831s |
| Trajectory B | 4 | Parallel to light source | Dynamic | Dawn/Dusk | Slow & Fast | SPAD & ZED2 | Off | 2.42m | 55s / 19s |
| Trajectory C | 4 | Toward light source | Dynamic | Dawn/Dusk | Slow & Fast | SPAD & ZED2 | Off | 2.57m | 58s / 13s |
| Trajectory D | 4 | Away from light source | Dynamic | Dawn/Dusk | Slow & Fast | SPAD & ZED2 | Off | 2.86m | 52s / 17s |
| Trajectory E | 4 | Spot turns | Dynamic | Dawn/Dusk | Slow & Fast | SPAD & ZED2 | Off | 0.53m | 58s / 15s |
| Trajectory F | 1 | Compound dynamic | Dynamic | Dawn/Dusk | Slow | ZED2 | Off | 11.51m | 309s |
| Trajectory G | 1 | Compound closed-loop | Dynamic | Dawn/Dusk | Slow | ZED2 | Off | 12.18m | 265s |

We also include wheel odometry, IMU, and ground truth data, all formatted as CSV text files to ease readability. Wheel odometry and inertial measurements are provided relative to the frames of reference depicted in Fig. 2.

**Wheel odometry data** consists of timestamps, computed linear and angular velocities, as well as the estimated 3-dimensional position derived from the averaged readings of every in-wheel encoder. Wheel odometry data were recorded at 25 Hz, with values expressed in (*m/s*) and (*rad/s*).

**IMU data** include timestamps, the rover's orientation—expressed as a unit quaternion $(x, y, z, w)$ with $w$ as the scalar component—, the angular velocity and linear acceleration cartesian vectors, and their associated covariance values. Inertial measurements were recorded at 60Hz, with values expressed in ($m/s^2$) and ($rad/s$).

**Ground truth** measurements are provided with respect to a global frame whose origin is defined according to Fig. 1. Note that in this global frame, the negative y-axis corresponds to the gravity-aligned vector. Ground truth data includes timestamps, the rover's 3-dimensional position, and its orientation—formatted in the same quaternion representation as the inertial measurements. Ground truth was recorded by the motion capture cameras at 25 Hz, with position values expressed in meters.

Alongside formatted data, we provide recorded raw *rosbags* for every trajectory in an MCAP format. Each *rosbag* includes ground truth data, wheel odometry and inertial measurements, and in those cases where the ZED2 stereo camera is used, both left and right rectified RGB and monochrome images.

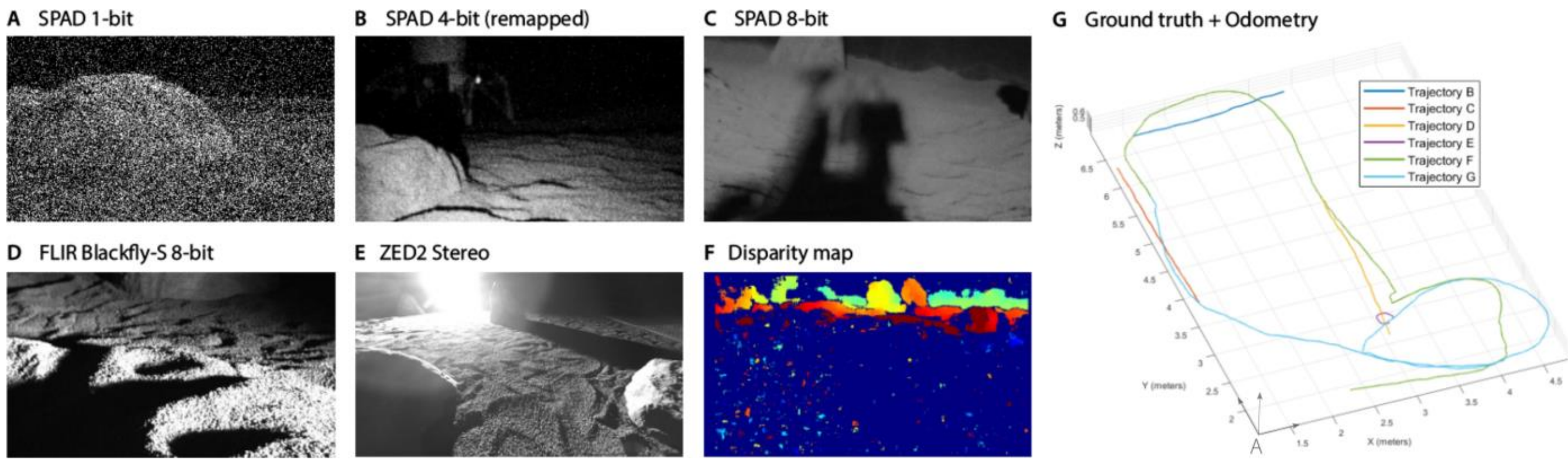

Fig. 4: Examples of the different images, formatting, and odometric measurements provided in the SPICE-HL$^3$ dataset[6]. Images are cropped to a 16:9 ratio in this figure for illustration purposes only.

*Data synchronization*

Data was recorded on multiple dedicated computers and at different frame rates depending on the data source. To ensure accurate timestamping and synchronization across all computers, we established an internal network run by a Network Time Protocol (NTP) server. This configuration allowed all the commanding computers to synchronize their clocks with millisecond accuracy. Timestamps are formatted in seconds since January 1, 1970, 00:00:00 (UTC), expressed with nanosecond-level precision; e.g.,`1726152291.870697076`.

*List of trajectories*

The recorded trajectories can be divided into 3 main groups. All the trajectories include wheel odometry, inertial, and ground truth measurements. Details of each trajectory are listed in Table III, while the ground truth trajectories are depicted in Fig. 4G.

**Trajectory A** consists of a long, compound path formed by 100 waypoints, covering a significant portion of the test field. Waypoints are spaced roughly 25 cm apart to ensure sufficient overlap between consecutive frames. At each waypoint, the rover halts to capture static images and record measurements, mimicking the operational behavior of most planetary rovers. Taking advantage of the stationary nature of this trajectory, we tested seven different illumination conditions at each waypoint (i.e., Reference, Noon, Dusk/Dawn, and Night, as defined in the Methods section), with the last three conditions both with and without the rover's headlights switched on. For each lighting scenario, images are taken at five different exposure times. During this trajectory, only monocular images were captured using the FLIR Blackfly-S and the SPAD512 cameras. For the SPAD camera, 256 binary frames (1-bit format) were recorded for each exposure time.

**Trajectories B–E** involve short movements, ranging from 12.5 to 62 s, during which images are continuously acquired. Each trajectory represents a different driving direction relative to the Sun-simulating spotlight (see details in Table III). For all these trajectories, the illumination scenario is set to Dawn/Dusk with the rover's headlights turned off. Images were captured using both the ZED2 stereo camera and the SPAD512 camera. For the SPAD512, single-photon frames were saved in a 4-bit format with fixed equivalent exposure times as follows:

- Trajectory B: 400 µs per frame at 215–510 Hz
- Trajectory C: 135 µs per frame at 120–440 Hz

- Trajectory D: 100 µs per frame at 125–154 Hz
- Trajectory E: 150 µs per frame at 30–120 Hz

These exposure times were chosen based on the recommended settings from the SPAD built-in autoexposure module. Note that for the SPAD512, exposures must be preset before image acquisition. The associated frame rates correspond to the slow and fast motion recording, respectively. The dataset includes raw 4-bit images. However, a script to remap these images into an 8-bit colormap for visualization purposes has been included in the code provided alongside this dataset (see Code Availability). In contrast, the ZED2 camera was operated in autoexposure mode, and we provide both left and right 8-bit rectified monochromatic images. Image sequences from both cameras were recorded at two different rover speeds: slow (5 cm/s and 0.1 rad/s) and fast (50 cm/s and 1 rad/s).

**Trajectories F–G** consist of continuous, compound paths where stereo images are continuously acquired using only the ZED2 camera. The rover moves across the test field for a distance similar to that of the static *trajectory A*, but this time combining different segments of motion without stopping at each step. *Trajectory G* forms a loop by revisiting some of the locations previously traversed during *trajectory F*. We chose to separate these into two distinct trajectories due to a significant gap in the timestamps between them. However, users of the dataset have the option to treat both trajectories as a single sequence by merging the imaging data.

*Data structure*

Data files are provided in individual ZIP files, including every sequence for each of the recorded trajectories. These ZIP files are structured as illustrated in Fig. 5. Every image file follows the same naming convention, i.e., ‘`cameratype_timestamp_frameID.*`’.

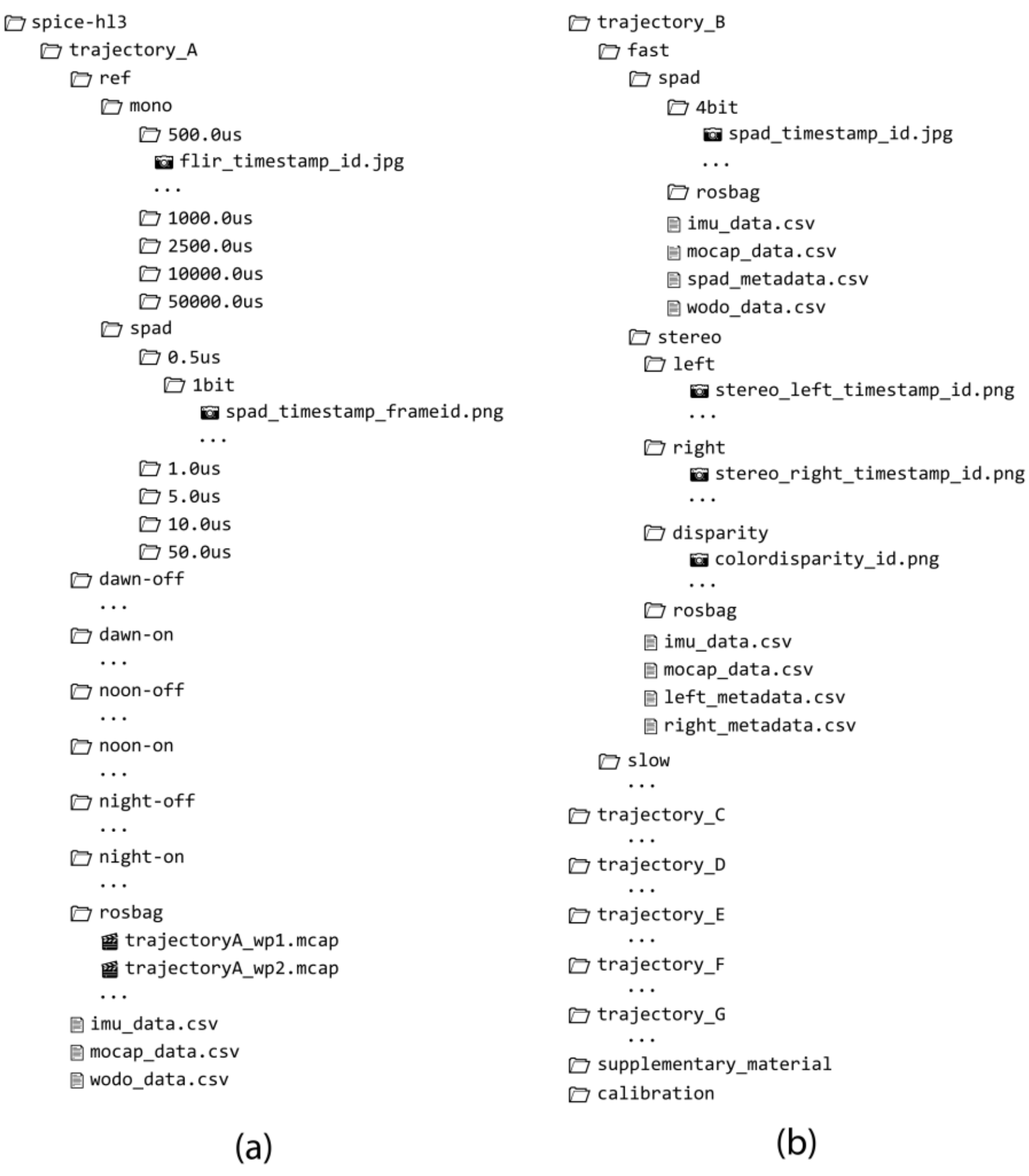

Fig. 5: Structure of the provided ZIP files storing all the trajectories and data file naming convention. Two types of directory structures can be found: a) a directory tree corresponding to trajectory A and b) a directory tree corresponding to trajectories B–G. For each type of directory, we illustrate an example structure of only one of the trajectories. The remaining trajectories listed for each type follow a similar structure.

### *Sensor calibration*

The dataset includes the intrinsic calibration parameters for each of the cameras used. This can be found within the `camera_intrinsics.json` file. Camera intrinsics have been estimated using a 2-coefficient calibration and the pinhole camera model, which include the focal lengths ($f_x$, $f_y$), the optical center ($c_x$, $c_y$), and the radial and tangential distortion parameters ($k_1$, $k_2$, $p_1$, $p_2$). We provide the original set of images used for calibration in the event that the end-user of the dataset prefers a different method. Note that single-photon calibration images are provided in 8-bit format.

To evaluate the localization accuracy of visual odometry and SLAM algorithms, the precise spatiotemporal relationship between cameras and ground truth measurements must be known. We implemented the target-based calibration approach described in [23]. This method uses offline, full-batch nonlinear least squares optimization to determine camera calibration parameters, namely: ${}^{C}_{M}T$, which represents the rigid body transformation from the global frame, $M$, to the camera frame, $C$; and $t_d$, which defines the time offset between the ground truth and camera clocks. Camera-to-global and camera-to-camera extrinsics, as well as all the transformation matrices from IMU and wheel encoders, are listed in the `sensor_extrinsics.json` file. We also provide the original set of calibration images used. Additionally, we provide the CAD model of each rover as part of this dataset.

## TECHNICAL VALIDATION

During and after the recording of the dataset, we performed a number of checks to validate the integrity, consistency, and utility of the data.

### *Data synchronization*

The different computers commanding each of the cameras and the one logging ground truth data were synchronized through an NTP network. Despite providing millisecond accuracy, for long data recording sequences, this approach could lead to compounded offsets between the timestamps of the different data sources. Upon evaluation, we concluded that the impact of this drift was negligible given the duration of each trajectory in the dataset, the speed of the rovers, and the high frame rates used.

Simultaneously, we observed slight mismatches between the left and right image streams of the stereo camera. This problem may have been caused by a ROS2 time tagging issue or by the lack of hardware-driven strict synchronization between the two camera streams, leading to discrepancies in frame timestamps and occasional frame drops. To ensure reliable stereo correspondence, we filtered the dataset to retain only matched frame pairs and discarded any frames with unmatched timestamps. For this, we used the script `stereo_file_matching.m` provided in the associated GitHub repository of this dataset (see Code Availability). The original unfiltered frames can be accessed by exporting the raw data from the *rosbags* provided for each trajectory.

### *Delayed ROS2 pipeline*

During the recording of stereo camera data via ROS2, we observed that a subset of captured camera frames (approximately 8%) were assigned delayed timestamps relative to their capture sequence ID. We believe this issue might stem from the ROS2 image pipeline publication-time tagging, leading to occasional inconsistencies due to processing or queuing delays. As the frames themselves are sequentially numbered without loss, we addressed this by post-processing the data. Delayed timestamps are corrected and frames are renamed by interpolating expected capture times based on the average frame interval within each sequence. This ensures the consistent temporal alignment required for downstream tasks. For this, we used the script `clean_delayed_frames.m` provided in the associated GitHub repository of this dataset (see Code Availability).

Delays were also observed in the inertial measurements recorded. Although IMU was configured to operate at 200 Hz, an analysis of the data reveals that, after accounting for delayed and missing data packets, the effective sampling rate across all trajectories falls to approximately 60 Hz.

### *SPAD image quality characterization*

Due to the limited number of published studies evaluating the passive imaging quality of large SPAD pixel arrays, we conducted a focused evaluation to verify the integrity and consistency of the recorded image data under extreme illumination conditions represented in the dataset. We performed a basic statistical quality assessment of the SPAD, FLIR, and ZED image streams. For each sensor and illumination regime, we analyzed differences in brightness, detail retention, and the presence of image artifacts, using metrics such as dynamic range, histogram distribution, and PIQE (Perceptual Image Quality Evaluator). PIQE is a no-reference quality metric that estimates image distortion by analyzing local variance across non-overlapping 16×16 pixel blocks, providing an overall assessment of the percentage of low-quality or distorted blocks, i.e., those characterized by extreme saturation, low contrast, or pronounced blurring[24]. PIQE is, therefore, inversely correlated with perceptual quality. These metrics were computed both per-frame and aggregated over complete trajectories, using the most visually challenging scenarios within our dataset (i.e., dawn/dusk, night, and sun-facing).

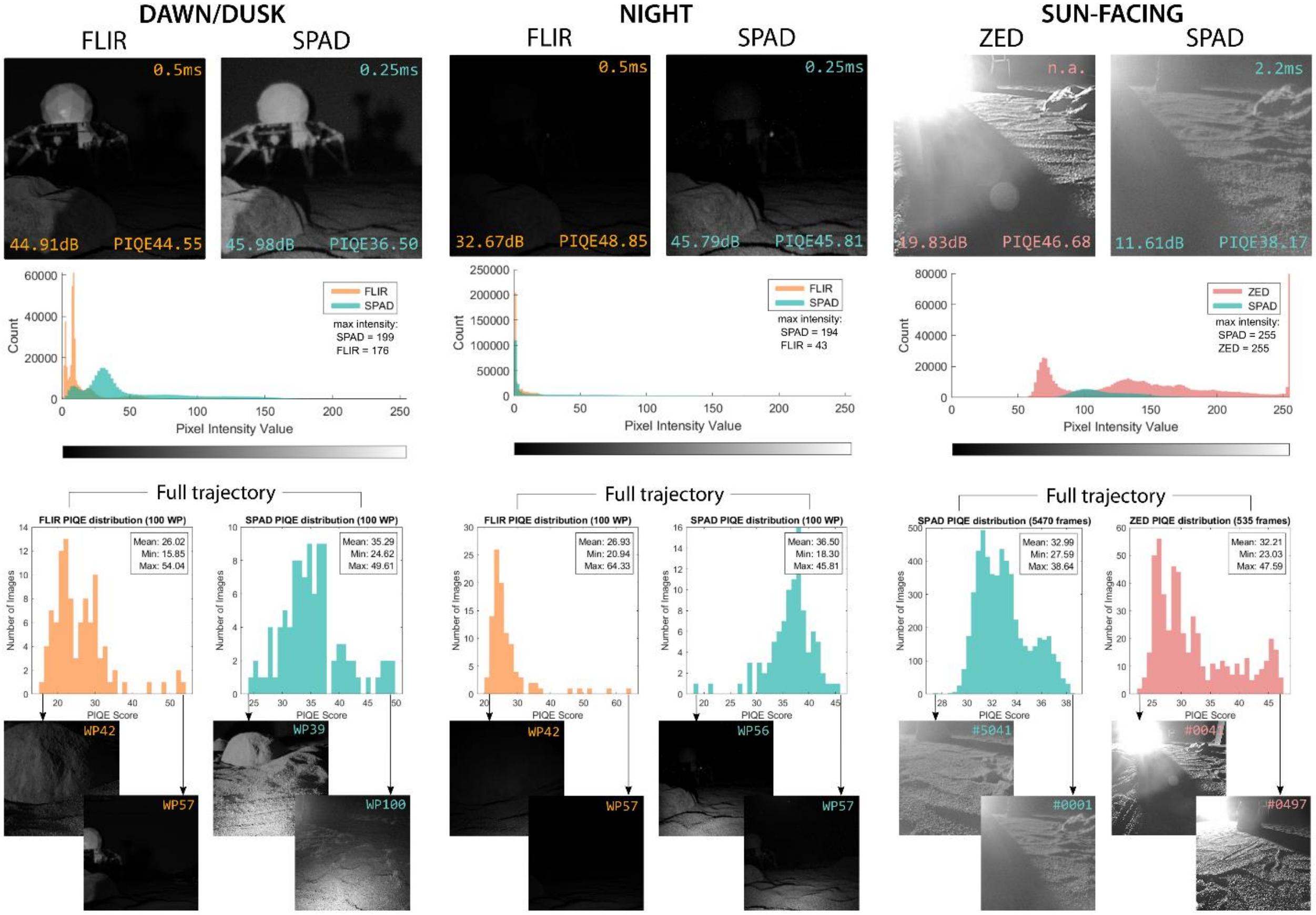

Fig. 6: Performance evaluation of the FLIR, ZED, and SPAD cameras under extreme visual conditions based on dynamic range, histogram distribution, and PIQE score. FLIR and ZED frames are cropped to match SPAD's field of view. SPAD frames are digitized to 8-bit images. DAWN/DUSK and NIGHT frames belong to trajectory A, while SUN-FACING frames belong to trajectory C. The top of the figure presents single frames from waypoint 58 for trajectory A and waypoint 10 for trajectory C. Histogram distributions are presented below the frames. The bottom of the image illustrates the distribution of PIQE scores for each camera and illumination condition over the complete trajectories. Best and worst PIQE-rated frames are displayed for context. Note: x-axis ranges differ across subplots and are adapted to the observed score ranges.

Fig. 6 illustrates representative frames and their corresponding histogram distributions, dynamic range, as well as the distribution of PIQE scores across sequences and sensors. The results confirm that the dataset spans a wide range of lighting conditions, from very low signal to high dynamic range scenes, while exhibiting stable sensor behavior and no evidence of systematic corruption, excessive frame dropouts, or acquisition failures. Outlier frames identified by extreme metric values were manually inspected and found to correspond to genuinely challenging but valid environmental conditions rather than data integrity issues.

Additional insights can be drawn from this evaluation. When comparing the performance across sensors, the single-photon camera delivers the most robust and uniform performance under extreme lighting, excelling in low-light and high-dynamic-range scenes by preserving structure and detail. Under these conditions, conventional cameras struggle with undersaturation and motion blur in low-light scenes or oversaturation when facing the Sun. More specifically, the SPAD512 achieves higher dynamic range, more uniform intensity distributions, and resolves image features even at half the 8-bit-equivalent exposure time, though with lower contrast in bright scenes. This is especially evident along the right-side edge of the rock located right in front of the rover in the dawn/dusk and night cases. The ZED2 camera provides strong contrast in well-lit conditions but loses critical details near light sources (see, e.g., the background rock in the upper-left corner of the image in the sun-facing scenario, only fully visible in the SPAD frame), while the FLIR camera performs well overall but degrades in very low light. As revealed in this complementary work[8], sun-facing drives represent one of the most challenging perceptual scenarios for vision-driven robots and autonomous vehicles due to the oversaturation and associated artifacts caused by direct sunlight and the limited ability of most cameras to simultaneously resolve both brightly lit and shadowed areas in a single shot.

It should be highlighted that metrics like PIQE, while useful for approximating perceptual quality, are sensitive to scene texture and lighting variability, and thus may not fully reflect the operational performance of a given sensor across diverse conditions. Extracting conclusions from single frames using texture-based metrics can be misleading, particularly in environments with significant lighting variation, as those experienced in our lunar analog trajectories. At the bottom of Fig. 6, we present the distribution of PIQE scores for each camera and illumination condition, computed over full trajectory sequences (i.e., 100 waypoints of trajectory A during dawn/dusk and night, and 5,470 SPAD 8-bit frames alongside 535 ZED2 frames from the sun-facing trajectory C). For each case, the best- and worst-rated frames based on PIQE scores are displayed to provide qualitative context. Interestingly, despite the SPAD's consistent ability to capture finer details and produce tighter, more stable PIQE distributions (clustered around the mean), both the FLIR and ZED2 cameras achieved lower absolute minimum and mean PIQE values due to their slightly higher resolution (texture-rich images) and narrower fields of view (less specular reflections).

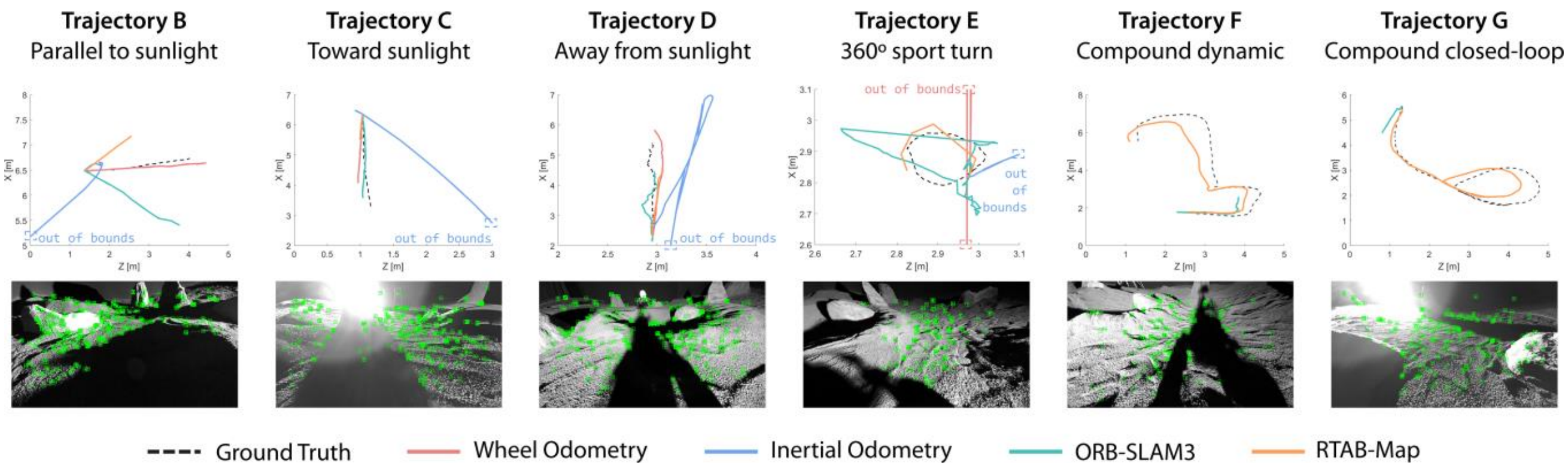

Fig. 7: Results of running wheel, inertial, RTAB-Map, and ORB-SLAM3 visual odometry methods over different trajectories in the dataset. For each trajectory, a representative critical frame is shown below the plot, corresponding either to the point of maximum absolute trajectory error (ATE) or to the point preceding a localization failure.

*Additional integrity and usability checks*

As an additional end-to-end data validation test, we verified that standard feature detection and matching, scene segmentation, and visual odometry pipelines can be executed on selected sequences without runtime errors or data compatibility issues. Preliminary results of this performance evaluation were reported in a separate publication[8]. The successful execution of these pipelines implicitly confirms that the camera calibration, timestamping, sensor synchronization, and data formatting are internally consistent and suitable for use with conventional computer vision pipelines for robotics. Additional details and a step-by-step guide on how to adapt the dataset to use well-known visual odometry and SLAM pipelines, such as RTAB-Map[25] and ORB-SLAM3[26], are included in the associated GitHub repository. Fig. 7 illustrates a summary of the results of running these relative localization methods using data from different trajectories.

## Usage Notes

During the development and processing of the SPICE-HL$^3$ dataset[6], we encountered several challenges and limitations that should be highlighted for future users of the data. These aspects primarily relate to the fidelity of the simulated lunar environment, the coordinate transformations between data sources, and the effect of secondary lights.

*Scene fidelity*

True lunar regolith composition and grain size distribution affect how light reflects off the surface. Lunar regolith contains grains orders of magnitude smaller than the smallest grain of soil used in the LunaLab facility (lunar regolith contains grains of 40–130 µm[27] compared to the 0.2–4-mm grains of the LunaLab soil). This mismatch affects the photometric characteristics of our images compared to those captured at facilities featuring advanced lunar regolith simulants meant to replicate lunar albedo to a higher degree of fidelity or those that will eventually be taken at the lunar poles. Some effects were irreproducible in the LunaLab, e.g., the opposition or halo-like effect caused by the coherent backscattering of light as it bounces off the crystalline minerals within the regolith grains when viewed from phase angles close to zero (i.e., backlighting). On the other side, while higher photometric accuracy would have been preferred, it would have also been extremely costly to do so. This is due to the size of the test bed and the safety measures required when operating with soils that include silty and clay particles. Nonetheless, the lower reflectivity of our soil compared to true regolith still serves us as a worst-case optical scenario.

In addition, the lighting setup within the LunaLab imperfectly replicates lunar sunlight conditions. Due to constraints related to the size of the test bed and the proximity of the light source, light decays inverse-squared within the length and width of the test bed. This leads to uneven illumination across the scene. Ideally, a perfectly collimated light source should be used so that illuminance remains uniform across the test bed, as it would be expected over relatively large distances on the Moon. Users of the data should expect slight inconsistencies in image exposure, particularly during the longer trajectories.

### *Coordinate transformations*

The default coordinate frame for collecting the ground truth data was defined so that the negative y-axis corresponded to the gravity-aligned vector, while the z-axis pointed toward the short side of the test field. This should be noted when comparing ground truth data with the motion estimated in either the optical frame or the robot frame.

The ZED2 stereo camera has a slight down tilt of 25º. On this rover, the ground truth tracker is mounted directly on top of the stereo camera rig and, therefore, is affected by this same pitch angle. In contrast, on the rover carrying the SPAD and FLIR cameras, the ground truth tracker is mounted flat on a separate mast behind the camera mounts.

### *Effect of secondary light sources*

Despite our efforts to recreate the low-scatter, high-contrast lighting conditions characteristic of lunar environments, we overlooked the impact of the motion capture system itself in artificially illuminating the scene. Specifically, the dim status LED rings on the motion capture cameras and the IR light they emit to track reflective markers on the rover inadvertently became secondary light sources. This is primarily noticeable under the most light-deprived conditions, such as during dusk/dawn and nighttime drive sequences without the use of headlights.

The faint visible light from the status LED rings is most apparent, despite affecting all cameras equally, in the longer-exposure images captured by the FLIR camera. This issue could have been mitigated by disabling or covering the status LED of all cameras. However, they served a useful purpose in identifying and troubleshooting motion capture camera issues during testing.

More significantly, the IR illumination used by the motion capture system for tracking the rovers introduced an unavoidable effect due to the partial sensitivity of most camera sensors to near-infrared light. This led to faint glows, ghosting, and occasional lens flares. These effects are evident in some of the ultra-fast binary frames captured by the SPAD. While these artifacts are likely present across all cameras, their prominence depends heavily on sensor characteristics and exposure times. Acknowledging the presence of these effects is important for correctly interpreting the dataset and guiding future sensor deployments under similar experimental setups. Future data collection will incorporate strategies to better isolate imaging sensors from unintended secondary light sources to further improve the fidelity of simulating lunar photometry.

## Acknowledgments

We would like to thank Pi Imaging Technologies for generously providing the SPAD512 camera used in the recording of this dataset.

## Data Availability

The dataset is publicly available at Zenodo[6].

## Code Availability

All code described in this paper can be accessed at `https://GitHub.com/spaceuma/spice-hl3`. We designed Python and Matlab scripts to be easily adaptable to the ultimate end-user needs. Further descriptions, requirements, and versions are stated in the git repository's `README.md`.

## Funding

This work was supported in part by armasuisse Science and Technology (contract number 8003538860) under the project "Monocular SPAD camera for enhanced vision in complex and uncertain environments." This work was partly conducted when the corresponding author was still affiliated with the Advanced Quantum Architecture Laboratory (AQUA) at EPFL, Switzerland. This work was also supported by the project INSIGHT (PID2024-160373OB-C21) funded by the Spanish Ministry of Science, Innovation, and Universities, the Spanish State Research Agency, and the European Regional Development Fund (ERDF, UE).

## Author Contributions

D.R.-M. conceived the experiments, formatted and prepared the dataset, built and wrote the code in the code repository, and drafted the original manuscript. D.R.-M., D.vdM., J.S., and A.B. refined the implementation of the experiments and prepared the rovers for data acquisition. D.vdM., J.S., A.B., and M.O.-M. arranged the experimental facility prior to the experiments. D.vdM. and A.B. operated the rovers and supervised the network. J.S. assisted and conducted sensor calibration. D.R.-M., D.vdM., and A.B. conducted the rover experiments. C.J.P.P. and M.A.O.-M. defined data formatting and advised on data validation. All authors reviewed, edited, and approved the final manuscript.

## Competing Interests

The authors declare no competing interests.